\documentclass[10pt,twocolumn,letterpaper]{article}

\usepackage{cvpr}

\definecolor{cvprblue}{rgb}{0.21,0.49,0.74}
\usepackage[pagebackref,breaklinks,colorlinks,allcolors=cvprblue]{hyperref}
\usepackage{xcolor}
\usepackage{float}
\usepackage{ulem}
\usepackage{siunitx} 
\usepackage{gensymb}

\def\paperID{5981} %
\def\confName{CVPR}
\def\confYear{2025}

\title{ArtiScene: Language-Driven Artistic 3D Scene Generation\\ Through Image Intermediary}

\author{
Zeqi Gu$^{1,2}$~~~~~~Yin Cui$^1$~~~~~~Zhaoshuo Li$^1$~~~~~~Fangyin Wei$^1$~~~~~~Yunhao Ge$^1$\\
Jinwei Gu$^1$~~~~~~Ming-Yu Liu$^1$~~~~~~Abe Davis$^2$~~~~~~Yifan Ding$^1$\\
$^1$NVIDIA~~~~~~$^2$Cornell University
}

\definecolor{mathbrace_color}{rgb}{0.2,0.5,1.0}

\newcommand{\beq}{\begin{equation}}
\newcommand{\eeq}{\end{equation}}

\begin{document}

\twocolumn[{
\renewcommand\twocolumn[1][]{##1}
\maketitle
\begin{center}
  \captionsetup{type=figure}
  \vspace{-8mm}
  \includegraphics[width=\linewidth]{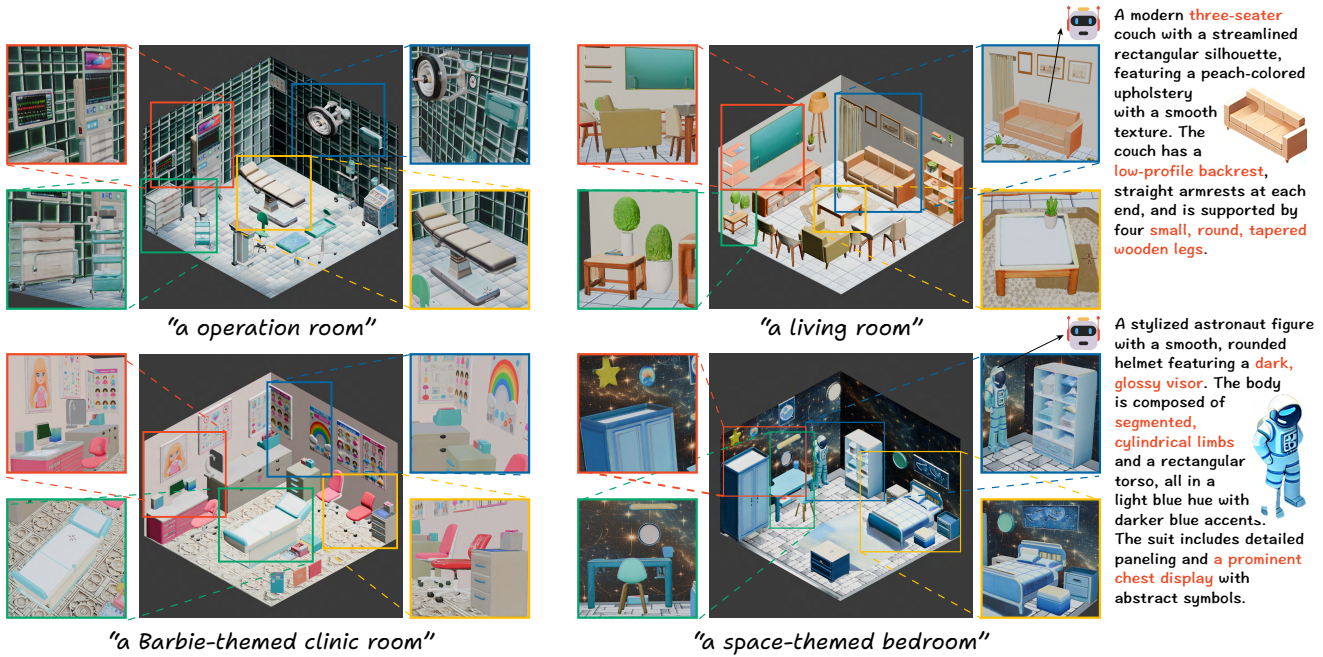}
  \captionof{figure}{
    We present ArtiScene, a training-free, language-driven 3D scene generation pipeline that can design diverse, aesthetic, and easily editable scenes across a wide range of categories and styles from a text prompt. The figure shows four results with zoom-in details from various angles. Everything is generated, including furniture, decorative objects, scene layouts, floors, and walls. We first leverage text-to-image models to generate a 2D image intermediary and then extract rich layout and style information from it. 
    For example, the rightmost column shows the object appearance and geometry acquired from the image via object segmentation and LLM-powered description.
  }
  \label{fig:qual_holodeck}
\end{center}
}]

\maketitle

\begin{abstract}
Designing 3D scenes is traditionally a challenging and laborious task that demands both artistic expertise and proficiency with complex software. 
Recent advances in text-to-3D generation have greatly simplified this process by letting users create scenes based on simple text descriptions. However, as these methods generally require extra training or in-context learning, their performance is often hindered by the limited availability of high-quality 3D data. In contrast, modern text-to-image models learned from web-scale images can generate scenes with diverse, reliable spatial layouts and consistent, visually appealing styles. Our key insight is that instead of learning directly from 3D scenes, we can leverage generated 2D images as an intermediary to guide 3D synthesis. In light of this, we introduce ArtiScene, a training-free automated pipeline for scene design that integrates the flexibility of free-form text-to-image generation with the diversity and reliability of 2D intermediary layouts.
First, we generate 2D images from a scene description, then extract the shape and appearance of objects to create 3D models. These models are assembled into the final scene using geometry, position, and pose information derived from the same intermediary image. Being generalizable to a wide range of scenes and styles, ArtiScene outperforms state-of-the-art benchmarks by a large margin in layout and aesthetic quality by quantitative metrics. It also averages a 74.89 $\%$ winning rate in extensive user studies and 95.07 $\%$ in GPT-4o evaluation. \href{https://artiscene-cvpr.github.io/}{Project page here.}
\end{abstract}
    
\section{Introduction}
\label{sec:intro}

Designing 3D scenes is a critical task with ample applications in embodied AI, virtual reality~\cite{fu2025anyhome}, indoor and outdoor design~\cite{feng2024layoutgpt, paschalidou2021atiss}, and automated space planning \cite{fu2024scene, huang2023embodied}. Yet completing such a task often requires human experts to spend a considerable amount of time on the manual steps involved in the process. Recent breakthroughs in generative artificial intelligence offer substantial potential to enhance the efficiency of current processes, fostering research efforts in this field~\cite{rombach2022high, ramesh2021zero, openai_chatgpt}. A popular strategy has been to condition the 3D generation on a user input text prompt describing their desired scene~\cite{chung2023luciddreamer, cohen2023set,po2024compositional,yang2024holodeck}. Such framing of the problem enables us to leverage text-conditioned generative models that have made impressive progress recently, such as diffusion models~\cite{ho2020denoising,song2020denoising} and Large Language models (LLMs)~\cite{devlin2019bert,vaswani2017attention}. However, existing solutions face two major problems: first, without fine-tuning or in-context learning with additional 3D training data, their ability to generalize to new scene categories or styles is constrained.
Second, many methods rely on asset retrieval to populate generated layouts with geometry, which restricts the contents of a scene to arrangements of assets from a limited pool of existing 3D models. 
These strategies are severely limited by the scarcity of high-quality 3D data, which has been a long-standing problem in computer vision.

In light of these challenges, we instead seek solutions from the 2D world. Trained on billions of 2D Internet images, text-to-image models have shown significant power in synthesizing high-quality images. We find that diffusion models~\cite{rombach2022high,ramesh2021zero} excel at generating layout images with diverse, reliable spatial arrangements and consistent, visually appealing styles. Therefore, we use these images as an intermediary from the input text to the output 3D scene. Through leveraging the extraction of rich layout information from 2D designs, we elevate the text into a 3D scene that aligns with the user's initial input. We first detect the furniture, decorations and small objects within the image intermediary, and prompt LLMs to describe their appearance and geometry. Then instead of asset retrieval from some data pool, we feed the segmented images and textual descriptions to a single-view 3D generation model to acquire customized 3D assets one by one. This design allows much more flexibility in the object appearance and geometry. Next, we deploy a single-view depth estimation model to estimate the position and dimension of each asset. Once the 3D assets are generated, we place them at the predicted positions and match their poses by rendering and comparing the rendered images with their 2D segment counterparts in the original guide image. After gathering all information on position, dimension, and pose, and generating texture images for the floor and walls (if exist), we assemble the generated 3D assets to form the final output scene.

In summary, our key contributions are three-fold:
\begin{itemize}
    \item We propose a novel approach to tackle the task of text-to-3D scene generation, which leverages 2D images as an intermediary to extract 3D layout information from them.
    \item We build our automated solution system, ArtiScene, which is capable of generating diverse, customizable, and modular 3D scenes with no extra training. 
    \item We conduct extensive quantitative and qualitative experiments to demonstrate the power of ArtiScene. With an object overlapping rate $6-10\times$ lower and a CLIP score higher than current state-of-the-arts by a large margin, our results are also consistently preferred by humans as shown in the user studies.
\end{itemize}

\section{Related Works}
\begin{figure*}[t]
  \centering
   \includegraphics[width=\linewidth]{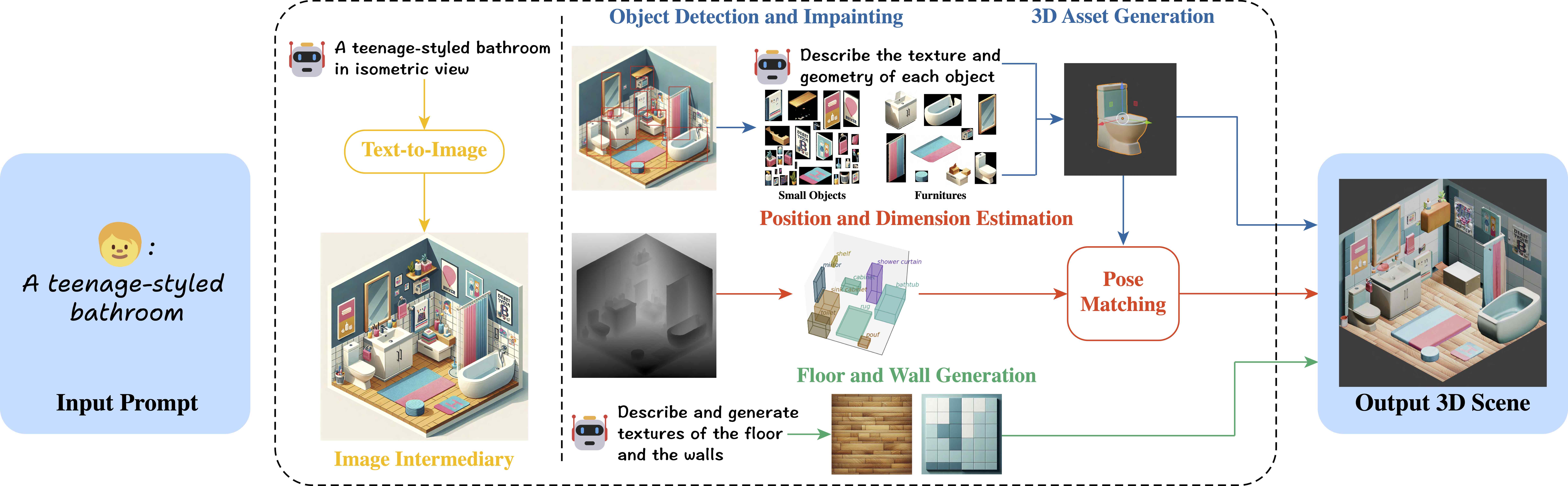}

   \caption{\textbf{Pipeline Overview.} Taking a text prompt as input, ArtiScene first prompts a text-to-image model for an image intermediary (yellow line). Then through object detection, inpainting, and prompting ChatGPT to describe the detected objects' appearance and geometry, we acquire a 3D model for each object (blue line). In parallel, we combine monocular depth estimation with the formerly detected 2D bounding boxes to estimate the 3D bounding boxes of each object (red line). We also synthesize the floor and wall textures for indoor scenes (green line). In the end, we assemble these acquired models and layout information to arrive at the final 3D scene.}
   \label{fig:pipeline}
   \vspace{-0.5cm}
\end{figure*}

There is a long track of research efforts to synthesize diverse and realistic scenes with generative networks. Earlier methods used feed-forward networks, variational autoencoders~\cite{gao2023scenehgn,yang2021scene} and generative adversarial networks to model the distribution of objects in indoor scenes. SceneFormer~\cite{wang2021sceneformer} introduced the use of transformers to add furniture to scenes, and later ATISS~\cite{paschalidou2021atiss} demonstrated that a single transformer model is sufficient to generate more realistic and efficient arrangements. However, these methods often require extensive training on specific datasets, limiting their ability to generalize to objects and scenes that are underrepresented in the training data. 

Another family of methods builds procedural synthesis pipelines~\cite{raistrick2023infinite,raistrick2024infinigen,zhou2024scenex}. Such method requires one tailored generator per object category, making extensions to novel scenes labor-intensive. For example, InfiniGen Indoors~\cite{raistrick2024infinigen} only targets at 5 room types, and cannot handle new scene types nor styles.

With the advent of foundation models, such as diffusion models and LLMs trained on internet-scale data, many research areas, including 3D scene generation, have been significantly impacted. These models encapsulate a vast amount of knowledge about various data distributions in the human world, and recent efforts have focused on leveraging this knowledge for downstream tasks.

\subsection{Scene Generation with Diffusion Models}
Diffusion models excel in image generation, providing realistic and contextually accurate inpainting by seamlessly filling in missing or corrupted areas of an image.
Using a diffusion model, many methods~\cite{nguyen2024housecrafter,hollein2023text2room,fridman2024scenescape,yu2024wonderjourney,yu2024wonderworld,hoellein2023text2room} inpaint or outpaint an entire scene and then lift it to 3D using depth estimation and reconstruction techniques. However, this strategy makes object-level editing cumbersome, and the results often exhibit multi-view inconsistencies. ControlRoom3D~\cite{schult2024controlroom3d} and Ctrl-Room~\cite{fang2023ctrl} address these issues by incorporating 3D geometric guidance. Still, since these approaches generate a single 3D model for the entire scene from a panorama image proxy, they lack modular composability. In contrast, our approach does not focus on precise reconstruction from images, but rather on extracting rich information to synthesize a coherent scene that adheres to the initial text prompt. We also use one independent mesh to represent each of the objects, and thus our result allows object-level modular editing.

Diffusion models can also provide gradient guidance for 3D generation through Score Distillation Sampling (SDS) loss. DreamFusion~\cite{poole2022dreamfusion} first proposed to optimize 3D NeRF~\cite{mildenhall2021nerf} representations by feeding rendered images to 2D diffusion models for supervision. Subsequent efforts aimed at generating more complex scenes learn compositional 3D representations with SDS ~\cite{po2024compositional,zhou2024gala3d,cohen2023set,bai2023componerf,wang2023luciddreaming,chen2024comboverse,fu2025anyhome}. Nonetheless, such optimizations are generally still limited to scenes with few objects, and the results often exhibit multi-view inconsistencies~\cite{lin2023towards, zhang2023scenewiz3d}.

\subsection{Scene Generation with LLMs}
Recently, people have found that LLMs possess strong symbolic reasoning skills. Since they are trained with program data, constructing layouts as structured programs allows them to ``imagine'' object locations from merely language tokens~\cite{feng2024layoutgpt,yang2024llplace,ocal2024sceneteller,fu2025anyhome,zhang2024scene,lin2024genusd}. However, relying solely on LLMs for scene generation often leads to physically implausible outcomes, such as overlapping objects. To mitigate this issue, some methods~\cite{yang2024scenecraft} introduce extra user controls such as bounding boxes, while others~\cite{yang2024holodeck,lin2024instructscene} optimize a scene graph under spatial relational constraints. However, the constraints remains mostly manually defined. As these approaches typically require extensive prompt engineering to prompt the LLM by in-context learning using highly relevant examples, the quality of generated results heavily depends on the examples. Additionally, 3D assets are retrieved from existing datasets based on programmatic output, which limits stylistic consistency. Even the largest current 3D asset datasets~\cite{deitke2023objaverse,deitke2024objaverse,deitke2022️} lack sufficient appearance granularity to ensure a coherent overall visual style.

In summary, compared to all aforementioned approaches, ArtiScene is unique in its balance between diversity, modularity and quality. As will be shown in Sec.~\ref{sec:results}, its strong performance in style and aesthetics makes it particularly useful to designers and artists.

\section{Method}
Given a condition $C$ describing a scene, our task is to generate a set of objects $O={o_1,o_2,...o_n}$ where each object $o_i$ consists of its 3D model $m_i$, size $s_i \in \mathbb{R}^3$, location $t_i \in \mathbb{R}^3$, and orientation $r_i \in \mathbb{R}^3$. 
The condition $C$ is a text prompt provided by the user. We begin by using a diffusion model to generate an image as an intermediary representation. 
Taking inspiration from architectural drawings, we generate an isometric perspective of the scene, which uses an orthographic camera rotated such that the projection of the y-axis remains vertical, and projections of the three major coordinate axes intersect at three equal angles. This standard perspective represents all three spatial dimensions of an object in a manner that is invariant to its placement in the scene. 
Once we have generated a guide image, the rest of our pipeline includes five steps, which we explain in the following subsections.

\begin{figure}[t]
  \centering
   \includegraphics[width=\linewidth]{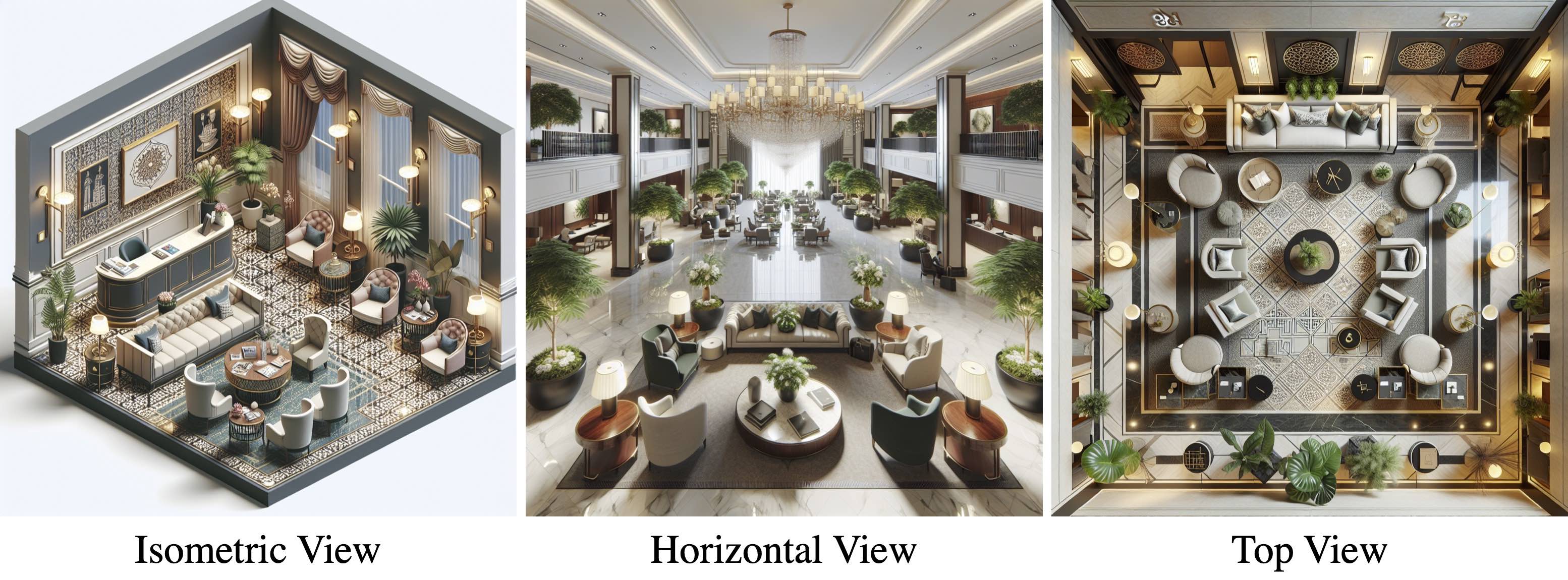}

   \caption{\textbf{Layout from Different Views.} Isometric projection conveys information about all three spatial dimensions of an object while making its appearance invariant to translation in the scene.
   }
   \vspace{-0.5cm}
   \label{fig:views_comp}
\end{figure}

\subsection{Object Detection and Inpainting}
The first step is to detect and segment the objects in the image. The target text labels that would be fed as input into a detection model are composed of two parts. Assuming the scene category (e.g., bedroom, restaurant, etc.) is known, we first query ChatGPT what furniture and decorative objects are common for that type of scene. Then we prompt ChatGPT to list out objects present in that specific proxy image. The two parts complement each other to provide an abundant name list of objects that are likely to be present. After the detection model outputs bounding boxes, we extract segmentation masks from each bounding box. Due to occlusions and imprecise segmentation, some objects will have holes, and we fill them in with plausible textures with an inpainting model. We keep track of detected furniture and decorative objects separately for later convenience (see Sec.~\ref{sec:post_processing}).

\subsection{Position and Dimension Estimation}
\label{sec:pose_est}
To project the 2D image to 3D, we first use a monocular depth estimation model to estimate the depth. 
We found that existing monocular depth models worked sufficiently well even on images with isometric perspective. 
Given the estimated depth $Z$, we project point $(x,y)\in \mathbb{R}^2$ to $(x',y',z')\in \mathbb{R}^3$ by:
\begin{align}
    x'=\frac{x-c_x}{f_x},\\
    y'=\frac{y-c_y}{f_y},\\
    z=\alpha*Z(x,y) \label{eq:z},
\end{align}
where $(c_x,c_y)$, $(f_x,f_y)$ are the center of the image and the focal length of imaginary camera respectively. This is a modification of the classical camera intrinsics projection formula, taking out the scaling effects of depth on $x'$ and $y'$. Additionally, as depth estimation model outputs an image in range 0 to 255 to represent the depth, we need to scale it to match with $x'$ and $y'$, so we multiply $z$ by $\alpha \in \mathbb{R}$ in Eq.~\ref{eq:z}. 

We transform the objects by $-45\degree$ vertically and $-35.26\degree$ horizontally to align with the world coordinates. In the end, we project the detected 2D bounding boxes to 3D to get their position and dimension information.

\subsection{3D Asset Generation}
\label{sec:asset_gen}
Our 3D generation model is conditioned on a text description and a single-view image of the object. We ask ChatGPT to describe each segmented image, focusing on the shape, geometry, and color. We ask it to avoid mentioning anything not related to the object itself, including but not limited to the viewpoint, the background, and especially remaining decorative objects and occlusion holes that are not resolved by the first step. Once we acquire the text, we feed it and the segmented image of the object into our generation model.

\subsection{Pose Estimation and Placement}
We follow the propose-select scheme widely adopted in pose estimation works. Assuming object $o_i$ only rotates around the z-axis, we render it at the predicted location $t_i$, rotating $\{0,45,90,135,180,225,270,315\}$ degrees, and use a feature matching network to find the pose that gives the result closest to the segmentation of that object, measured by L-2 distance in the feature space. Note that the object pose is correlated with its size, so we rescale its three dimensions every time to stay within the same predicted bounding box. In the end, we choose the pose $r_i$ and size $s_i$ that gives the lowest L-2 error, use them to rotate and resize the object $o_i$ respectively, and place the object at location $t_i$.

\subsection{Floor and Wall Generation}
To make the scene more realistic, we add plane meshes as floor and walls. We ask ChatGPT about the texture of the floor and the walls separately, and use its description to let DALLE-3~\cite{betker2023dalle3} generate wallpaper-styled images, which will be set as textures for the corresponding planes. The floor is horizontally placed at the lowest z value of all objects, centered at the mean $(x, y)$ coordinate of all objects, and the walls are vertically placed at the extreme values of all objects in the four horizontal directions (i.e. $\pm$x and $\pm$y). 

\subsection{Post-Processing}
\label{sec:post_processing}
The estimated positions of the furniture and small objects may contain errors. To minimize the influences of these errors, we first find the minimum and maximum values of all bounding boxes in the six-axis directions (i.e. $\pm$x, $\pm$y and $\pm$z). Then we iterate over all objects to check if any of their local extrema values is within a pre-defined distance margin to the scene extrema value in the same direction. If the check returns positive, we move that object to align with the extreme value of the entire scene. Intuitively, this step would push nearby objects to a wall or to the floor. It is common that some small objects are on top of some larger furniture pieces. To maintain their relative positions, we associate small objects with furniture if the top of a furniture bounding box is within a predefined distance to the bottom of a decoration object bounding box. Later on, if the furniture needs to be moved, we would move the associated small objects by the same amount. After the ``pushes'', we record the list of objects that are on the floor, and that on each of the walls respectively.

We further remove small occlusions caused by estimation errors. We sort all objects by their distance to the scene corner $(x_{min},y_{min},z_{min})$ in an increasing order. Starting with the object closest to it, we check pairwise occlusions between it and the remaining objects that are further from the corner. If there is an occlusion, we move the further object in one of the six directions that needs the smallest moving distance. If the object is recorded as being in contact with the floor or a wall in the previous step, the positive and negative directions orthogonal to the contact surface are excluded from consideration. For example, if an object has been recorded as being on the floor, it could only move in one of the $\pm$x, $\pm$y directions if there is an occlusion, such that it always remains on the floor.

\section{Implementation}
\label{sec:implementation}
We use version 4o for ChatGPT~\cite{openai_chatgpt4o_2024}, and DALLE-3~\cite{betker2023dalle3} as the diffusion model. We use GroundedDINO~\cite{liu2023grounding} for detection and segmentation. We observe the detection may miss some targets when there are many objects in an image, and thus we run the detection step twice. After the first run, we inpaint the holes caused by decorative objects with Remove Anything~\cite{yu2023inpaint}. Then we use the inpainted image as input to the second run, as it provides a clearer view of furniture that may have been partially occluded at first. To form the final detection and segmentation results, we combine the detected decorative objects from both runs. For furniture, we directly use the detection list from the second run, as they do not change in quantity or category after the first inpainting of decorative objects - the only change is that their appearances become more complete due to the removed and inpainted occlusions. 

We use Depth-Anything-2~\cite{depth_anything_v2} as the depth estimation model and find $\alpha=\frac{1}{300}$ to scale the predicted depth image works for most scenes. We call a public API~\cite{nvidia_edify_3d} for 3D object generation. For pose matching, we use a fusion of Stable Diffusion~\cite{rombach2022high} and DINO-v2~\cite{oquab2023dinov2} as the feature extractor for pose estimation. As the remaining incompleteness is mostly caused by occlusions between large-piece furniture, instead of between furniture and decorative objects, we switch to a more geometry-aware inpainting model, Pix2Gestalt~\cite{ozguroglu2024pix2gestalt}, to fill in the larger holes. As its results depend on seeding, we query ChatGPT to select the best inpainting result based on realism, appearance completeness, and texture consistency.

\section{Experiments}
\label{sec:results}
We compare with state-of-the-arts benchmarks that utilize foundation models for scene layout and style designing and possess object-level composability like our approach. In Sec.~\ref{sec:layout_eval}, we focus on comparing spatial arrangement to show that our image intermediaries indeed possess rich layout information. In Sec.~\ref{sec:overall_eval}, we compare end-to-end generation results to demonstrate that our superiority in visual quality and style aesthetics. We end this section by showing useful applications and extensions in Sec.~\ref{sec:applications}.

\subsection{Layout Evaluation}
\label{sec:layout_eval}
For 3D scene layout generation, we compare with LayoutGPT~\cite{feng2024layoutgpt}. As it does not consider design aesthetics, we follow its original evaluation scheme by evaluating on the same scene categories (i.e. bedrooms and living rooms) without specifying styles. We generate 42 results for each room type with our method, and use all corresponding test results from LayoutGPT for comparison. 

\paragraph{Quantitative Evaluation.} As our method does not require floor sizes as an input condition, Object Overlaping Rate (OOR) is a more suitable metrics than Out-Of-Bound (OOB) in this case. We exclude object categories in our results that are not considered in LayoutGPT. For both methods, if the bounding boxes of one chair and one desk overlap, we do not count it, as this tends to indicate that the chair is simply inserted into the desk. As shown in Table.~\ref{tab:OOR}, benefiting from the direct visual cues from the proxy image, our efficient 3D estimation strategy and de-occlusion post-processing, our method gives more than 6$\times$ OOR reduction for bedrooms and 10$\times$ for living rooms.

\begin{table}
  \centering
  \begin{tabular}{lSS}
    \toprule
    Method & OOR $\downarrow$ & Avg. \#Furn \\
    \midrule
    LayoutGPT (B)& 37.26 &4.30\\
    Ours (B)& \textbf{6.48}&6.97\\
    \midrule
    LayoutGPT (L)& 27.77&6.23 \\
    Ours (L)& \textbf{2.19}&8.66\\
    \bottomrule
  \end{tabular}
  \caption{\textbf{Quantitative Comparison w/ LayoutGPT.} (B) indicates
the bedroom scene subset, and (L) represents the living room
subset. 
Our method tends to predict a denser layout, while staying low in overlap rate, ``Avg. \#Furn'' denotes Average Number of Furniture in the room, indicating our generated scenes are richer in contents.
}
  \label{tab:OOR}
  \vspace{-0.3cm}
\end{table}
\begin{figure*}[h]
  \centering
   \includegraphics[width=\linewidth]{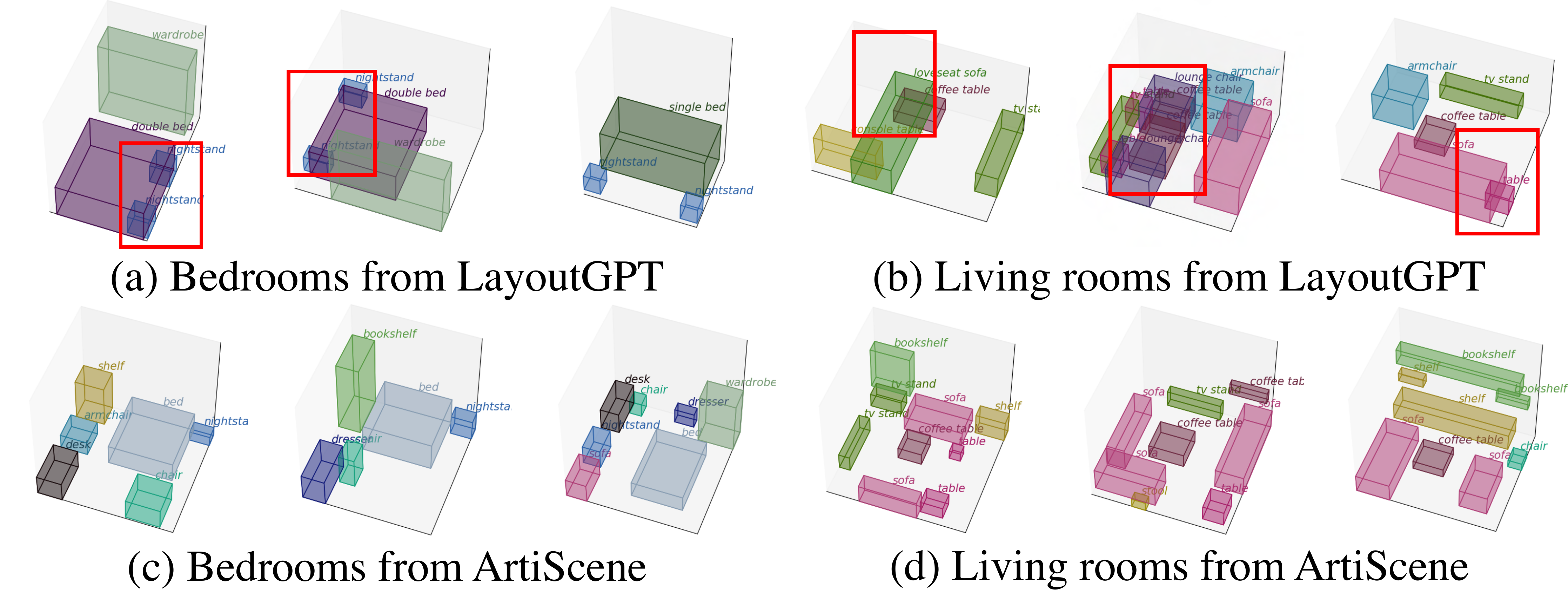}

   \caption{\textbf{Qualitative Comparison w/ LayoutGPT.} We visualize layout predictions as 3D bounding boxes from a side view rendering. Each furniture type is presented by a fixed color for all plots. The regions marked by \textcolor{red}{\textbf{red boxes}} contain severe overlaps, which occurs much more often in LayoutGPT results. As LayoutGPT needs in-context learning from these datasets, their results tend to memorize the training distribution and is sometimes overly simple (e.g. row (a)). Our results are much more diverse and reasonably sophisticated. Please zoom-in to see more details.}
   \label{fig:layout_comp}
\end{figure*}
\vspace{-8pt}

\paragraph{User Study.}
While OOR can be a useful metric, it fails to capture many other important aspects of good layout. 
Therefore, we further validate layout plausibility through user study. Our user studies were conducted on the Cloud Research platform with 60 participants. For bedrooms and living rooms respectively, we sampled and paired up 15 results from ArtiScene and LayoutGPT, and invited 30 participants to answer two questions for each pair. Question 1 asks about \textit{Layout Plausibility}: how logically and naturally the furniture and decor are spatially arranged. Question 2 asks for their preference considering not only the spatial arrangement, but also the category, diversity and quantity of furniture. After evaluating all 15 pairs, the participants need to provide their overall preference among the two groups of samples. As shown in Table~\ref{tab:layoutGPT_user_study}, the percentage of population that prefer our results is consistently around or above two-third, for all questions and all scene categories.

\begin{table}
  \centering
  \begin{tabular}{l *{3}{S[table-format=2.2]}}
    \toprule
    Method & {Layout $\uparrow$} & {Preference$\uparrow$} & {Overall$\uparrow$} \\
    \midrule
    LayoutGPT (B) & 33.55 & 31.19 & 27.42 \\
    Ours (B) & \textbf{66.45} & \textbf{68.81} & \textbf{72.58} \\
    \midrule
    LayoutGPT (L) & 31.19 & 24.95 & 12.91 \\
    Ours (L) & \textbf{68.81} & \textbf{75.05} & \textbf{87.09} \\
    \bottomrule
  \end{tabular}
  \caption{\textbf{User Study Comparison w/ LayoutGPT.} (B) indicates the bedroom scene subset, and (L) represents the living room subset. The Layout column shows results for layout plausibility (Question 1), Preference represents user preference considering furniture quantity and diversity (Question 2), and Overall provides an overall rating of the two sample groups (Question 3).}
  \label{tab:layoutGPT_user_study}
  \vspace{-0.4cm}
\end{table}

\paragraph{Qualitative Evaluation.}

We provide visualizations of the predicted layouts for bedrooms and living rooms in Fig.~\ref{fig:layout_comp}. LayoutGPT results seems to memorize patterns that reflect training bias, such as always placing two nightstands on each side of a bed symmetrically for bedrooms. Corresponding to the above evaluations, our results contain less overlap, and appear more diverse and natural. 

\subsection{Scene Generation Evaluation}
\label{sec:overall_eval}
Next, we compare end-to-end generation results with Holodeck, the benchmark that, to our knowledge, is the closest to our method in terms of problem framing and overarching strategy. We demonstrate the superiority and robustness of our method by evaluating on a wide range of scenes. We use 8 scene categories from MIT Indoor Scenes dataset~\cite{quattoni2009recognizing}, covering not only common domestic scenes (bedroom, bathroom, dining room, etc.), but also public and commercial ones such as meeting rooms, locker rooms, waiting rooms, etc. For the aesthetic aspect, we ask ChatGPT to acquire 9 common decoration styles: Victorian, modern, art deco, farmhouse, coastal, Bohemian, Asian zen, Chinese and teenager. We further increase the diversity with 9 more themes of pop culture and arts: Barbie, Star War, panda, space, stellar, sport, forest, Van Gogh and Monet, and 3 themes of randomly picked pure color: aqua, purple and pink. We randomly combine themes and styles with room types, with the prompt template ``a X-styled/themed Y'', where X is the name of the style or theme, and Y is the room type. 

We use DALLE-3~\cite{betker2023dalle3} to generate 3 images for each prompt, and filter out those that have CLIP score below 30, to ensure the initial intermediary guide has high quality. Our final test sets include 111 samples, and we generate Holodeck results with the same text prompt distribution. 
\begin{figure*}[t]
  \centering
   \includegraphics[width=0.9\linewidth]{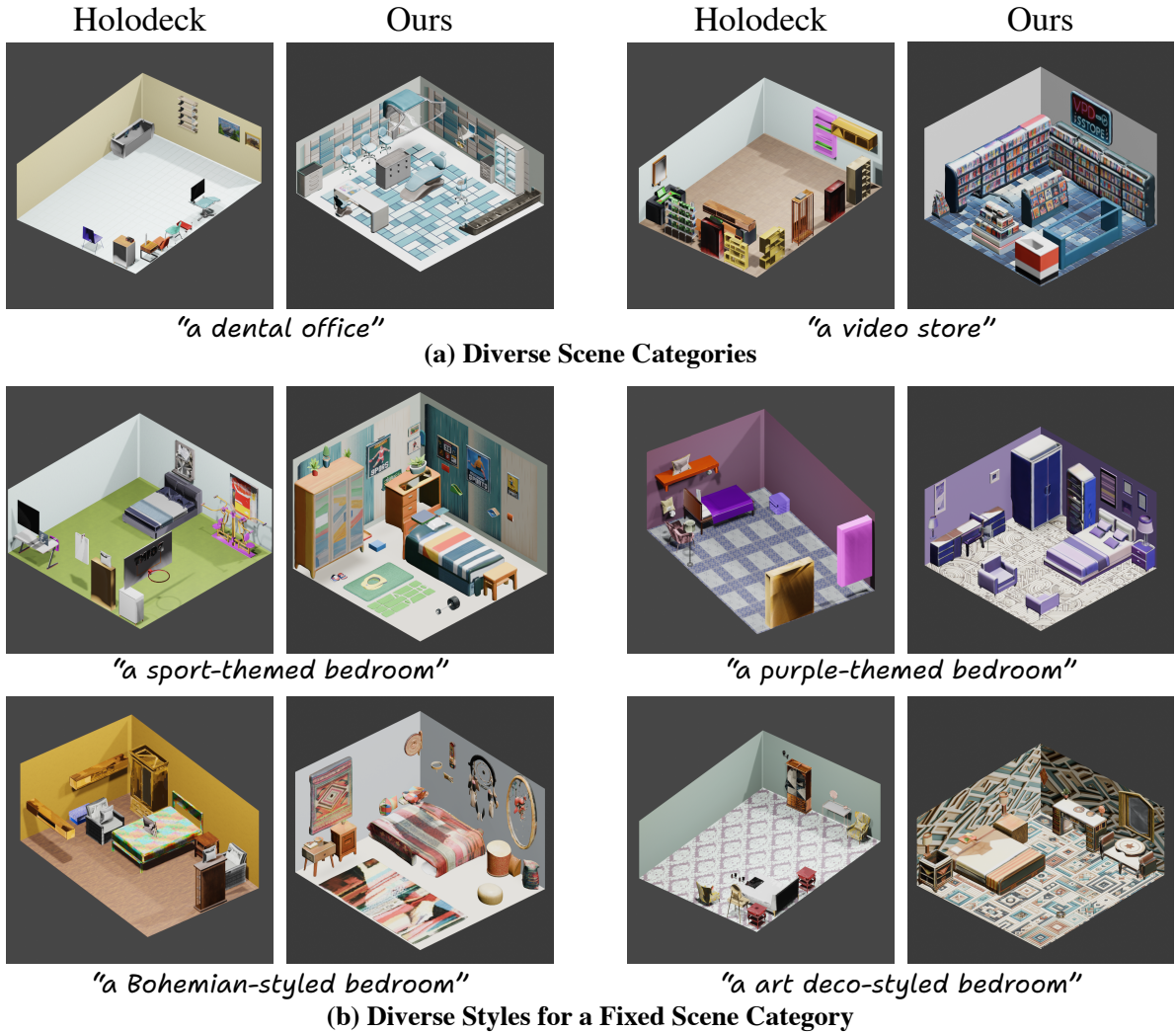}

   \caption{\textbf{Qualitative Comparison w/ Holodeck.} In (a) we show more scene categories sampled from the MIT dataset, and in (b) one fixed category (bedroom) with different styles and themes that we used for our evaluation. Our method is robust across a wide range of scene categories and aesthetics. Please zoom-in for more details, and please refer to our supplemental for more results.}
   \label{fig:qual_holodeck}
\end{figure*}

\paragraph{Quantitative Comparison.}
\begin{table*}
  \centering
  \begin{tabular}{l *{6}{S[table-format=2.2]}}
    \toprule
    Method & {CLIP$\uparrow$} & {Functionality$\uparrow$} & {Style$\uparrow$} & {Aesthetic$\uparrow$} & {Layout$\uparrow$} & {Overall$\uparrow$} \\
    \midrule
    Holodeck & 26.73 & 11.84 & 1.83 & 1.83 & 4.53 & 4.53 \\
    Ours & \textbf{29.45} & \textbf{88.15} & \textbf{98.16} & \textbf{98.16} & \textbf{95.46} & \textbf{95.46} \\
    \bottomrule
  \end{tabular}
  \caption{\textbf{Quantitative Comparison w/ Holodeck.} Our CLIP score and GPT-4o winning rates are higher across all metrics ($\%$).}
  \label{tab:CLIP_GPT_Holodeck}
\end{table*}
\vspace{-0.5cm}

In Table~\ref{tab:CLIP_GPT_Holodeck} we report CLIP score~\cite{radford2021learning} to evaluate the alignment of the generated scene style with input prompts. For a more detailed analysis, we apply GPT-4o to compare rendered images of paired results from us and Holodeck. Each pair is generated using the same text prompt and randomly matched. We use four criteria: \textit{Layout Plausibility}, which is the same as in the user study against LayoutGPT, \textit{Room Functionality}: how much the chosen furniture fit the basic functionality of the target scene type; \textit{Style Consistency}: how closely the generated result aligns with the requested style, and \textit{Style Aesthetic}: the overall aesthetics of the images. We ask GPT-4o to think out loud and evaluate each aspect step-by-step, and in the end, make a conclusive choice of the better one. Our winning rate is overwhelming in every aspect.

\paragraph{User Study.}
In the same pairwise manner, we invited 75 participants to compare 39 pairs of scenes randomly selected from the generated results. The samples are evenly divided into 3 groups, and we assign 25 participants to each group. We use the \textit{Style Consistency} and \textit{Layout Plausibility} as criterion, with the same aformentioned definitions. For each pair, we add a third question about overall preference, which is analogous to the ending question in the user study against LayoutGPT and the step-by-step reasoning used for GPT-4o evaluation. Table~\ref{tab:user_study_holodeck} indicates our results are preferred by the vast majority in all aspects.

\begin{table}
  \centering
  \begin{tabular}{l *{3}{S[table-format=2.2]}}
    \toprule
    Method & {Style$\uparrow$} & {Layout$\uparrow$} & {Overall$\uparrow$} \\
    \midrule
    Holodeck & 17.03 & 27.41 & 20.26 \\
    Ours & \textbf{82.96} & \textbf{72.58} & \textbf{79.73} \\
    \bottomrule
  \end{tabular}
  \caption{\textbf{User Study Comparison w/ Holodeck.}}
  \label{tab:user_study_holodeck}
\end{table}
\paragraph{Qualitative Evaluation.}

In Fig.~\ref{fig:qual_holodeck}, we compare qualitatively with Holodeck. Benefiting from the high-quality guidance of the image proxies, our results are consistently more visually appealing, and as we successfully detect and faithfully generate most of the objects in the image, our scene has much richer content. For walls and floors, Holodeck uses a fixed texture library, whereas we generate texture images with diffusion models during runtime, which also adds great variation and elevates the overall atmosphere of the scene.  
\subsection{Extensions}
\label{sec:applications}
\begin{figure}[h!]
  \centering
   \includegraphics[width=\linewidth]{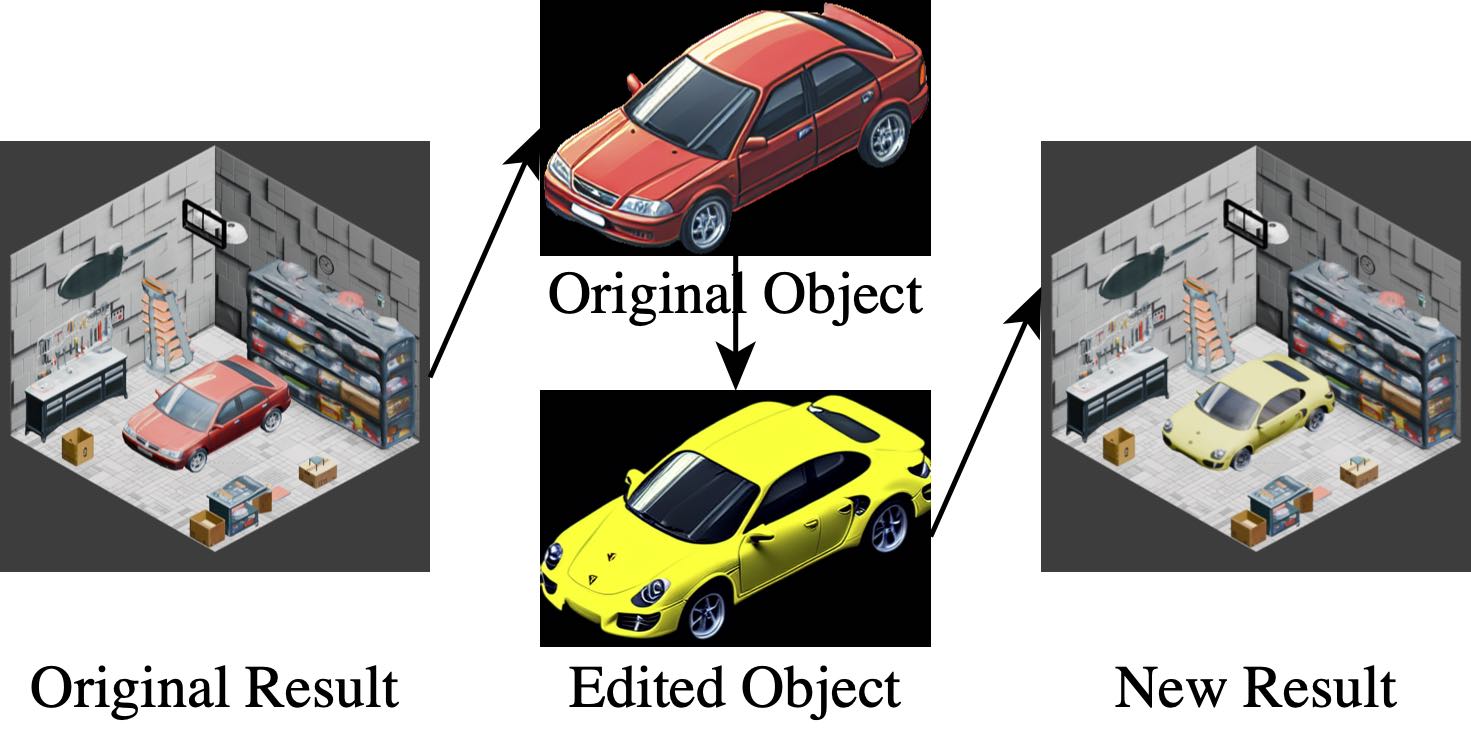}

   \caption{\textbf{Object Editing.} As ArtiScene generates a scene by generating individual 3D objects and then assembling, it allows fast and simple modular editing.}
   \label{fig:edit}
\end{figure}

An important advantage of our methods is its composibility and editability. As all objects are generated separately, we can change the appearance of one object without influencing the others. In Fig.~\ref{fig:edit}, we take the segmented car image that was produced when generating our original result on the left, and use Instruct-Pix2Pix~\cite{brooks2023instructpix2pix} to turn it into a red Porsche. We then rerun the steps in Sec.~\ref{sec:asset_gen} on that object, and put it into the scene using the 3D information from the original car.
\begin{figure}[h!]
  \centering
   \includegraphics[width=\linewidth]{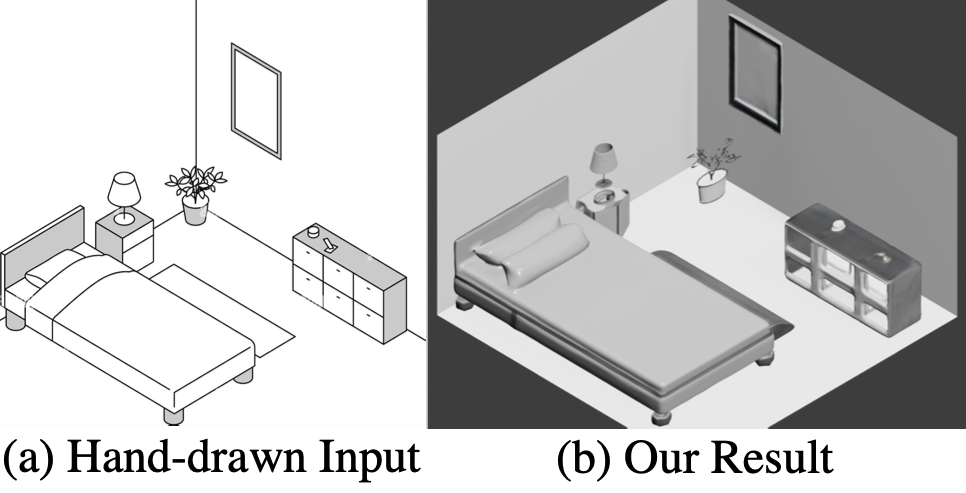}

   \caption{\textbf{Generation with Images Created by Human.} ArtiScene could be used with human-designed image input, skipping the initial text2im stage.}
   \label{fig:hand_drawn}
   \vspace{-0.5cm}
\end{figure}

As we make minimal assumptions about the target scene category, ArtiScene generalizes to outdoor scenes after small modifications. While the fact that our pipeline is built upon multiple model priors adds some complexity, it also offers useful modularity and opportunities for control. For example, our method could also skip the initial text prompt input, and be directly conditioned on an isometric layout image. Moreover, it generalizes well to layout images generated through other means, such as generated by professional software or drawn by a human (see Fig.~\ref{fig:hand_drawn} for an example), thus is able to incorporate human prior knowledge. Please refer to our supplemental for more results, as well as the ablation studies.

\section{Limitations and Discussions}
ArtiScene explores a novel way to leverage image models for 3D scene generation. However, some 3D understanding is traded for the advantage of learning from abundant 2D data. Although diffusion models generate high-quality layout images, so far it is weak at following more detailed text prompts, such as the specifications of object quantity and location. We also find the quality of generated isometric image is the scene is rarely plotted in isometric in real life, which often happens with more complicated, large-scale scenes such as museums, hospitals, etc. Generation with complex interactions and at high resolutions are active research areas, but orthogonal to our goals here. However, we emphasize that our key contribution is the overarching framework and strategy, and we do not rely on any specific model. It is very likely for stronger models to emerge in the future, and we believe substituting them into our pipeline would improve the result accordingly.
\vspace{-5pt}

\section{Conclusion}
We propose a novel text-to-3D indoor scene generation method that uses 2D image as an intermediary. Through extensive evaluations, we have demonstrated the superiority of our results in diversity, visual quality and physical plausibility. ArtiScene paves the way for various applications such as complex 3D interior design and immersive augmented and virtual reality.

\clearpage
{
    \small
    \bibliographystyle{ieeenat_fullname}
    \bibliography{main}
}
\clearpage
\setcounter{page}{1}
\maketitlesupplementary

\section{Ablation}
\subsection{Target Labels for Detection}

In Sec. 4 we mentioned we repeat the step of object detection twice: the first time after detecting furniture and small objects, we inpaint away the small objects, and detect for furniture and remaining small objects again. Thus, the detected furniture from the two runs stays the same amount, yet the detected small objects of the second round do not intersect with the detected set of first round, which have been removed by inpainting. We show the necessity of running twice in Table~\ref{tab:detected_small_objects}: counting the number of small objects in 184 generated scenes, more than 15$\%$ of them was detected in the second run. 
\subsection{Post-Processing}

Our de-occlusion post-processing is very effective in removing overlaps and placing the objects in a nearby reasonable position. As shown in Table~\ref{tab:ablation_OOR}, without our de-occlusion post-processing, the object overlapping rate increases. However, it will still be notably lower than the previous state-of-the-art LayoutGPT~\cite{feng2024layoutgpt}.
\subsection{Pix2Gestalt Inpainting and ChatGPT Selection}
After object detection, GROUNDING-DINO~\cite{liu2023grounding} performs segmentation within the detection 2D bounding box to generate more accurate masks. Such masks may contain holes that are due to occlusions by small objects, are lose a significant part of the object due to larger occlusions. We find Remove Anything~\cite{yu2023inpaint} no longer works for the latter, as the prior of that model seems to be stitching up the areas with smooth textures that could blend with its surroundings. Here we need such textures, as well as preservation of the general contour implied by the occluded version of the object. Inputting the inverse of the segmentation mask as the inpainting mask, We found Dall-E3~\cite{betker2023dalle3} often fill up the background too aggressively that the geometry of the foreground object is changed. The inpainted content of the Stable Diffusion family depend heavily on the shape of the inpainting mask: for example, if the region for inpainting has a square shape, SDXL~\cite{podell2023sdxl} tends to synthesize a square object instead of recovering the texture of our target shelf. After trying multiple combinations of different models and inpainting mask formulation, we found Pix2Gestalt~\cite{ozguroglu2024pix2gestalt} is the best at recovering the original geometry and texture of the target foreground object.
\begin{table}
  \centering
  \begin{tabular}{lcc}
    \toprule
    {First Run (Ratio $\%$)} &  {Second Run(Ratio $\%$)}  \\
    \midrule
    3118 (84.68)& 564 (15.31) \\
    \bottomrule
  \end{tabular}
  \caption{\textbf{\# Detected Small Objects.} Out of the two runs of object detection, 15.31$\%$ of all decor or small objects were detected in the second run.}
  \label{tab:detected_small_objects}
  \end{table}

\begin{table}
  \centering
  \begin{tabular}{l *{3}{S[table-format=2.2]}}
    \toprule
    Scene & {Ours (Full)} & {Ours (De-occlusion)} & {LayoutGPT} \\
    \midrule
    B & 6.48 & 27.30 & 37.26 \\
    L & 2.19 & 18.95 & 27.77 \\
    \bottomrule
  \end{tabular}
  \caption{\textbf{OOR Comparison.} B and L refer to bedroom and living room, respectively. Without our de-occlusion post-processing, the object overlapping rate increases (middle column), yet still notably lower than the previous state-of-the-art (right column). We do not consider margin when calculating overlaps.}
  \label{tab:ablation_OOR}
\end{table}
\begin{figure}[h!]
  \centering
   \includegraphics[width=\linewidth]{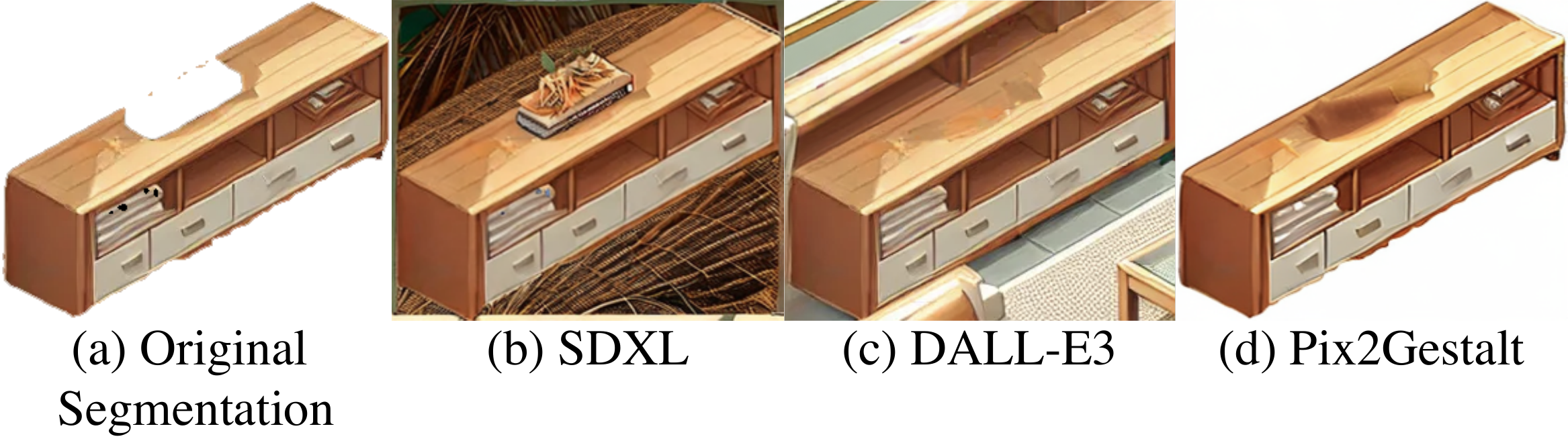}

   \caption{\textbf{Different Inpainting Methods.} We found Pix2Gestalt has strong geometry awareness. This object is the television shelf in Fig.~\ref{fig:Foundation_pose}.}
   \label{fig:edit}
   \vspace{-0.5cm}
\end{figure}

\begin{figure}[h!]
  \centering
   \includegraphics[width=\linewidth]{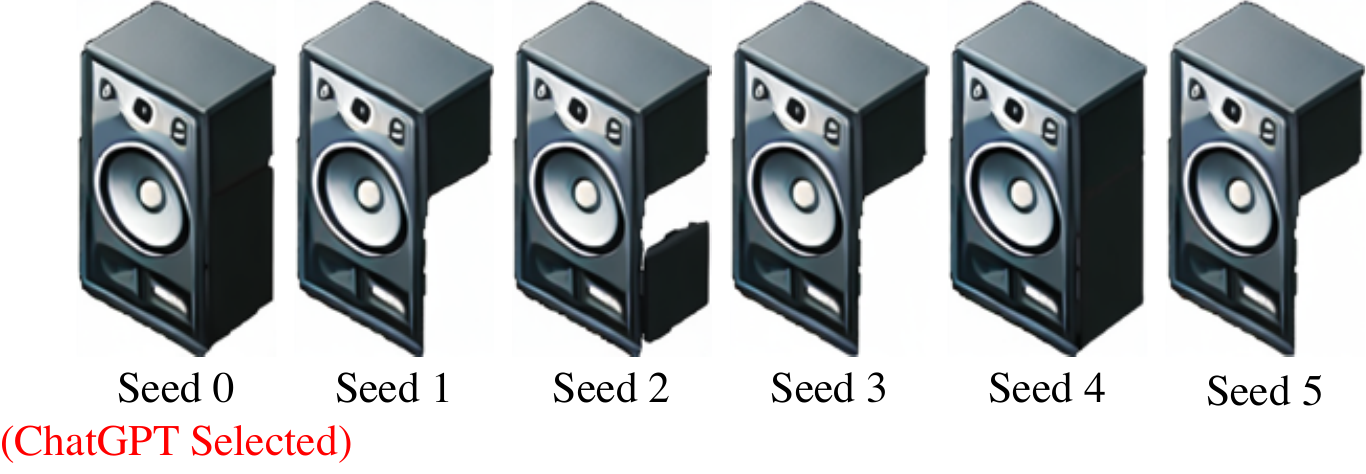}

   \caption{\textbf{Pix2Gestalt and ChatGPT Selection.} We use Pix2Gestalt to inpaint an incomplete audio. From seed 0 to 5, the results with seed 0 and 4 are more complete, ChatGPT successfully chose seed 0 as an satisfactory input for our following image conditioned 3D asset generation.}
   \label{fig:pix2gestalt_ChatGPT}
\end{figure}

One caveat is that the output of Pix2Gestalt varies with seeds. As shown in Fig.~\ref{fig:pix2gestalt_ChatGPT}, some results are more reasonable than others. To automate the selection of the best result, we prompt ChatGPT to select based on 3 aspects: (1) Realistic Object: how realistic it looks like the furniture category; (2) Complete Appearance: how complete the geometry and appearance is. If an image still holes or large occlusions on the object, it should have a lower score. And (3) Consistent Texture: how consistent the texture is. If the texture of one region is unrealistically inconsistent with its neighboring textures, such as a black spot, that is probably resulted from a failed inpainting, and such that image should have a lower score. Fig.~\ref{fig:pix2gestalt_ChatGPT} shows ChatGPT succeeds at picking one of the best results.

\subsection{3D Generation Conditioning}
\begin{figure}[h!]
  \centering
   \includegraphics[width=\linewidth]{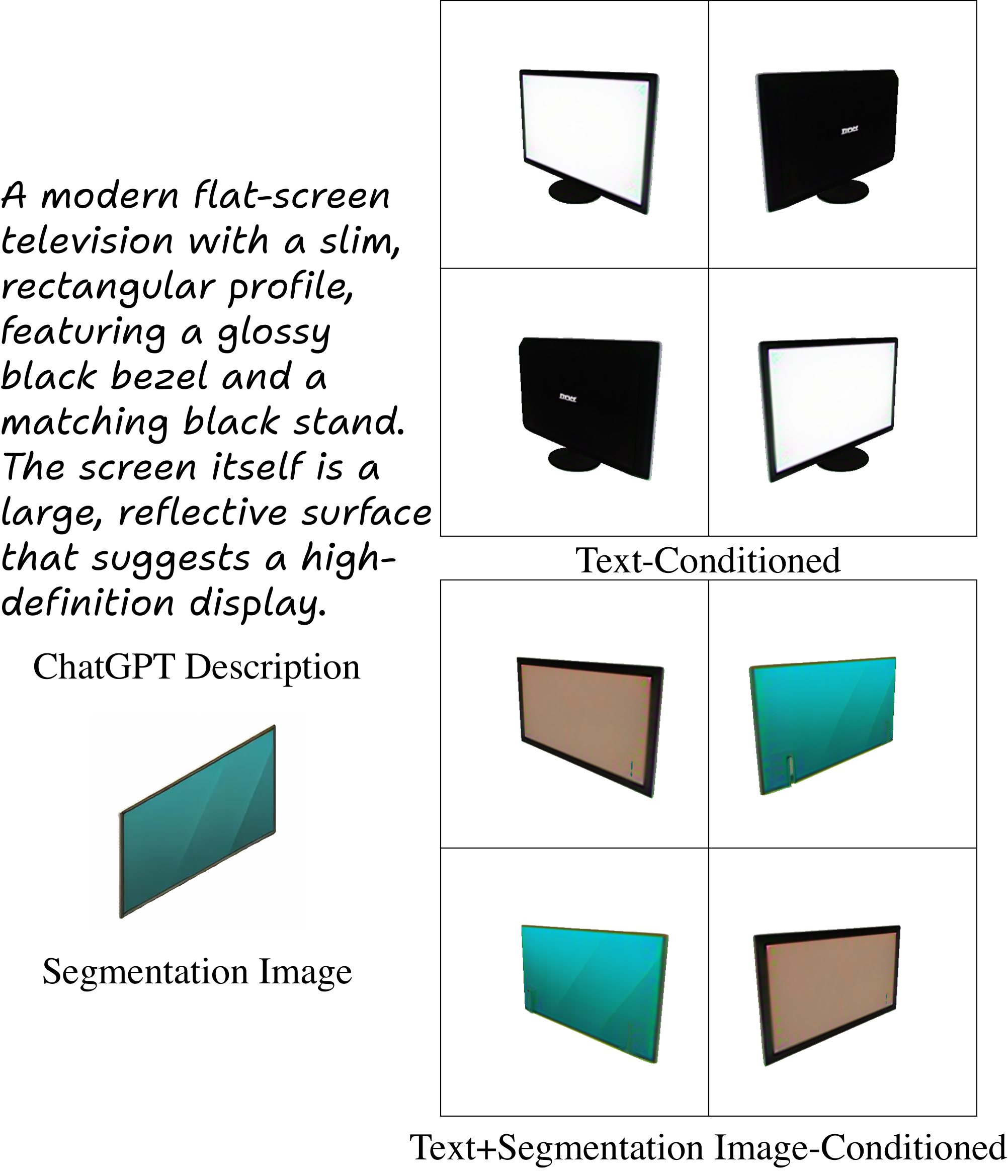}

   \caption{\textbf{Conditioning on Text and Image vs. Text Only.}  This object is the television in Fig.~\ref{fig:Foundation_pose}. We offer four perspectives of the 3D television model generated with text only (top right) and text plus the segmentation from the image intermediary (bottom right). With only text as inputs, the asset clearly adheres less to the original appearance in the image intermediary.}
   \label{fig:3d_gen_conditioning}
   \vspace{-0.5cm}
\end{figure}
We show the importance of generating 3D assest conditioning on both text and segmentation image in Fig.~\ref{fig:3d_gen_conditioning}.
\subsection{Pose Estimation}
\begin{figure}[h!]
  \centering
   \includegraphics[width=\linewidth]{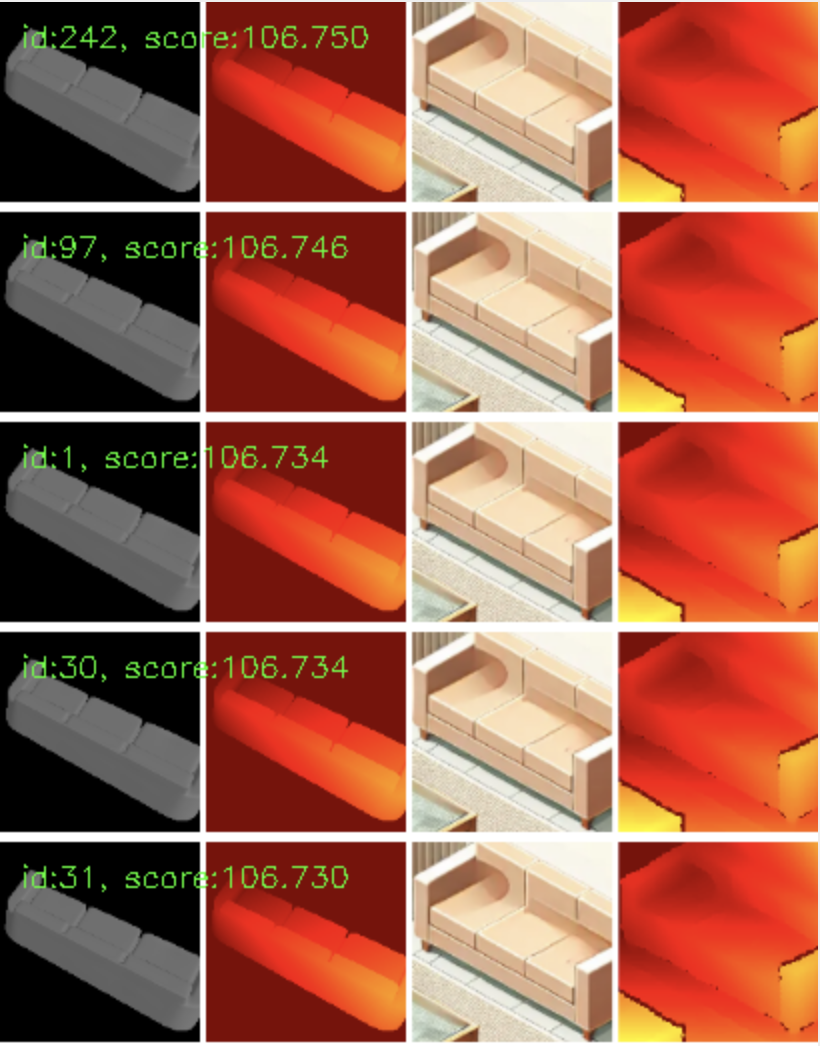}
\includegraphics[width=\linewidth]{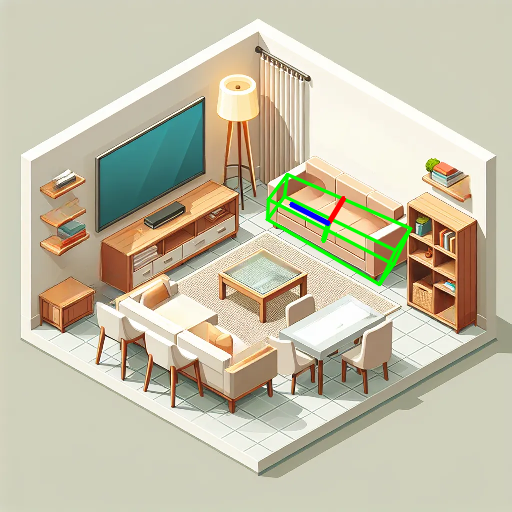}
   \caption{\textbf{Top 5 Pose Candidates by FoundationPose~\cite{ornek2025foundpose}.} The scores in green texts indicate the model confidence in each candidate pose. As we do not have ground truth camera parameters, and our 3D model is not exactly the same as the 2D counterpart in the image intermediary, even the top candidates are unsatisfactory. The green box with axes in the bottom figure is a visualization of the estimated pose with the highest score.}
   \label{fig:Foundation_pose}
   \vspace{-0.5cm}
\end{figure}
For pose estimation, we tried a state-of-the-art method focusing on real life robotic tasks, FoundationPose~\cite{ornek2025foundpose}. These methods often assume the 3D model of the target object is available, and could be rendered at different poses with known camera parameters. These rendered images would then be compared with the object seen in real life, and the pose of the closest image becomes the estimated pose output. Such approaches have two gaps from our use case: first, our image intermediaries are not rendered by cameras with common intrinsics, instead, they are isometric. Second, the 3D model generated at hand is not the same one as in the intermediary. As seen in Fig.~\ref{fig:Foundation_pose}, these errors accumulate and influence the final judgment of the model.
\section{Repetition Detection}
\begin{figure}[h!]
  \centering
   \includegraphics[width=\linewidth]{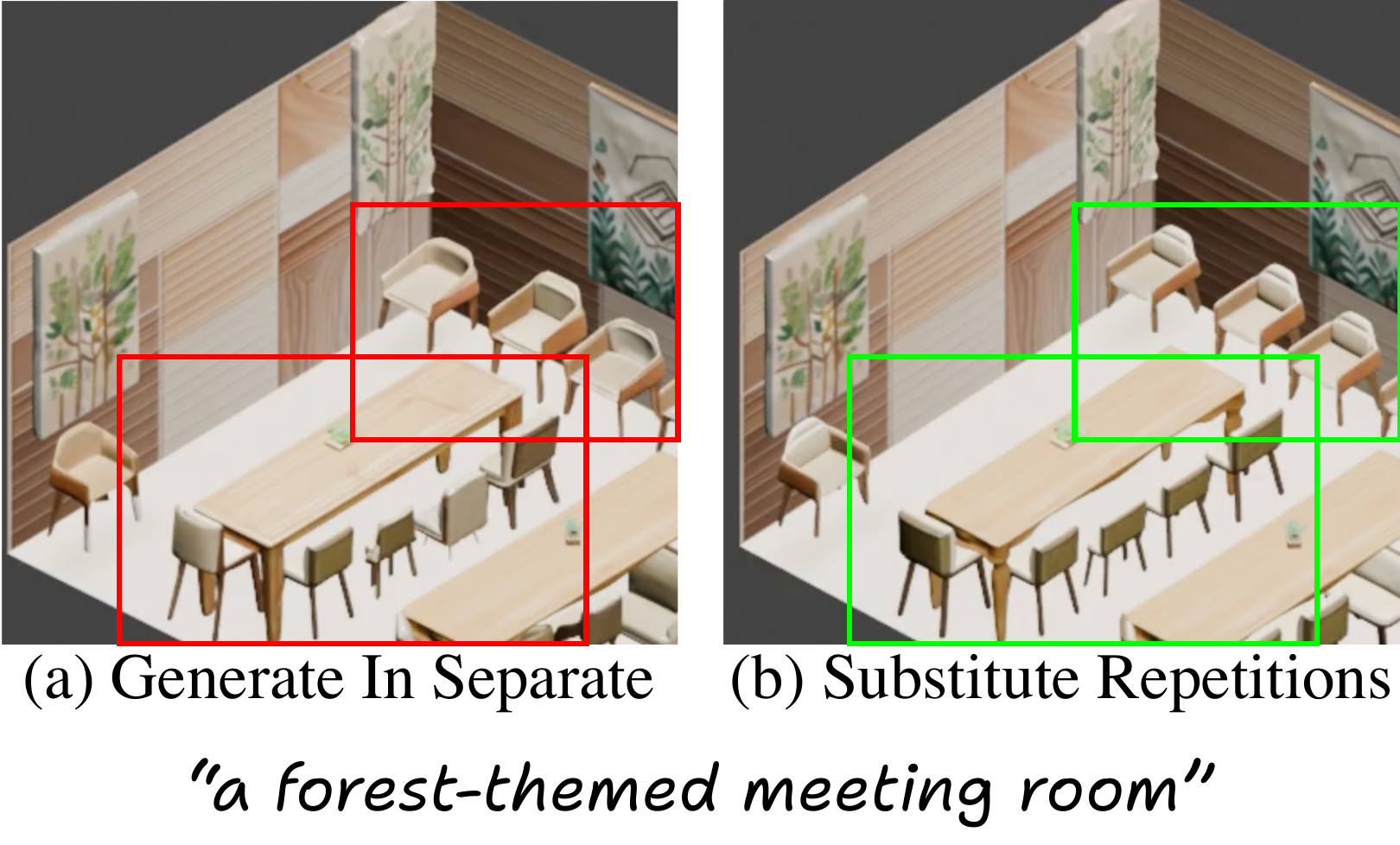}
   \caption{\textbf{Effects of Repetition Detection.} We replace the different chairs and armchairs in (a) with only one chair model and one armchair model, and keep the other parameters (pose, dimensions, positions) the same to create (b). The changed parts are highlighted by the rectangles.
   After applying the same 3D model for all similar objects, the result looks more uniform and realistic.}
   \label{fig:repetitions}
\end{figure}
To speed up the generation and make the final scene more consistent, we devise an optional module that automatically detects repetitions of objects. For public scenes such as classrooms and meeting rooms, it is common to have several pieces of furniture of the same model (e.g. chairs and desks). However, as they may be placed in different poses, we find the textual features more reliable than geometric features in determining if two pieces are of the same model. For a pair of objects $o_i, o_j$ We calculate the cosine similarities of the features of $T_i$ and $T_j$ extracted by SBERT~\cite{reimers2019sentence}. They are determined to be repetitions if the score is above 0.95. This default value is on the conservative side as we prefer less, correct substitutions over more but wrong ones. However, users could easily adjust it to achieve different level of uniformity. For example for Fig.~\ref{fig:repetitions}, we threshold at 0.89 to substitute more aggressively. We ask ChatGPT to pick the segmented image with the highest quality, and generate one 3D asset from it. Then for all repetitions we use this same asset for pose estimation and the final placement.

\section{More Discussion on Limitation}
Our automated pipeline generates 3D scenes in batches, yet scene-specific manual adjustments could further improve the results. For a very small subset of examples shown in this paper, we manually excluded 3D assets from the final scene if its quality is very low, and have manually selected certain Pix2Gestalt inpainted results over the GPT suggested ones. Choosing the hyperparameters more automatically and tailored to individual scenes would further improve the results, which we will leave as a future work.

Our method is slower than retrieval-based methods as we generate each object on the fly. We believe the speed could be improved as better component models emerge. After estimating layout and appearance information for each object from the image intermediary, we could also follow the practice of previous works to do 3D asset retrieval based on these features, instead of generation. We leave this interesting extension as another future work.

\end{document}